\documentclass[10pt,twocolumn,letterpaper]{article}

\usepackage{iccv}
\usepackage{times}
\usepackage{epsfig}
\usepackage{graphicx}
\usepackage{amsmath}
\usepackage{amssymb}
\usepackage{bm}
\usepackage{arydshln}
\usepackage{booktabs}
\usepackage{listings}
\usepackage{multirow}
\usepackage[table,xcdraw]{xcolor}
\usepackage{wrapfig}

\usepackage[pagebackref=true,breaklinks=true,letterpaper=true,colorlinks,bookmarks=false]{hyperref}

\iccvfinalcopy 

\ificcvfinal\fi

\begin{document}

\title{Bi-ViT: Pushing the Limit of Vision Transformer Quantization}

\author{Yanjing~Li\textsuperscript{1}, Sheng~Xu\textsuperscript{1}, Mingbao~Lin\textsuperscript{2}, 
Xianbin~Cao\textsuperscript{1},
Chuanjian~Liu\textsuperscript{4},
Xiao~Sun\textsuperscript{5},
Baochang~Zhang\textsuperscript{1,3}\\
\textsuperscript{1} Beihang University \quad
\textsuperscript{2} Tencent\quad
\textsuperscript{3} Zhongguancun Laboratory\\
\textsuperscript{4} Huawei Noah's Ark Lab \quad
\textsuperscript{5} Shanghai AI Laboratory \\
{\tt\small yanjingli@buaa.edu.cn}
}

\maketitle
\ificcvfinal\thispagestyle{empty}\fi

\begin{abstract}    

Vision transformers (ViTs) quantization offers a promising prospect to facilitate deploying large pre-trained networks on resource-limited devices.  Fully-binarized ViTs (Bi-ViT) that pushes the quantization of ViTs to its limit remain largely unexplored and a very challenging task yet, due to their  unacceptable performance. Through extensive empirical analyses, we  identify the severe drop in ViT binarization is caused by attention distortion in self-attention, which technically stems from the gradient vanishing and ranking disorder. To address these issues, we first introduce a learnable scaling factor to reactivate the vanished gradients and illustrate its effectiveness through theoretical and experimental analyses. We then propose a ranking-aware distillation method  to rectify the disordered ranking in a teacher-student framework. Bi-ViT achieves significant improvements over popular DeiT and Swin backbones in terms of Top-1 accuracy and FLOPs. For example, with DeiT-Tiny and Swin-Tiny, our method significantly outperforms baselines by 22.1\% and 21.4\% respectively, while  61.5$\times$ and 56.1$\times$ theoretical acceleration  in terms of FLOPs compared with real-valued counterparts on ImageNet. 

\end{abstract}

\section{Introduction}
\label{sec:intro}
%
%

Transformers, which have gained far-flung fame in natural language processing (NLP) area~\cite{devlin2018bert,qin2022bibert}, are also attracting increasing attention in lots of computer vision (CV) tasks, such as object detection~\cite{carion2020end}, image classification~\cite{dosovitskiy2020image} and many others~\cite{he2022masked,tian2022integrally}, impelling the widespread research on vision transformers (ViTs).
There has a natural fit for ViTs to achieve better performance simply by training a larger model on a larger data set.
For example, historical records show better performance of a ViT-H model~\cite{dosovitskiy2020image} accompanying with astonishing 632M parameters and 162G FLOPs. Such a high model complexity poses a great challenge to deploy models on platforms with short resource supplies. Therefore, both academia and industry call for an ultimate compression of these large models, and the past years have witnessed some promising techniques such as  network pruning~\cite{yang2021nvit,chen2023general}, low-rank decomposition~\cite{denil2013predicting}, knowledge distillation~\cite{jia2021efficient}, and quantization~\cite{li2022q,li2022qwacv}.

\begin{figure}[t]
    \centering
    \includegraphics[scale=.25]{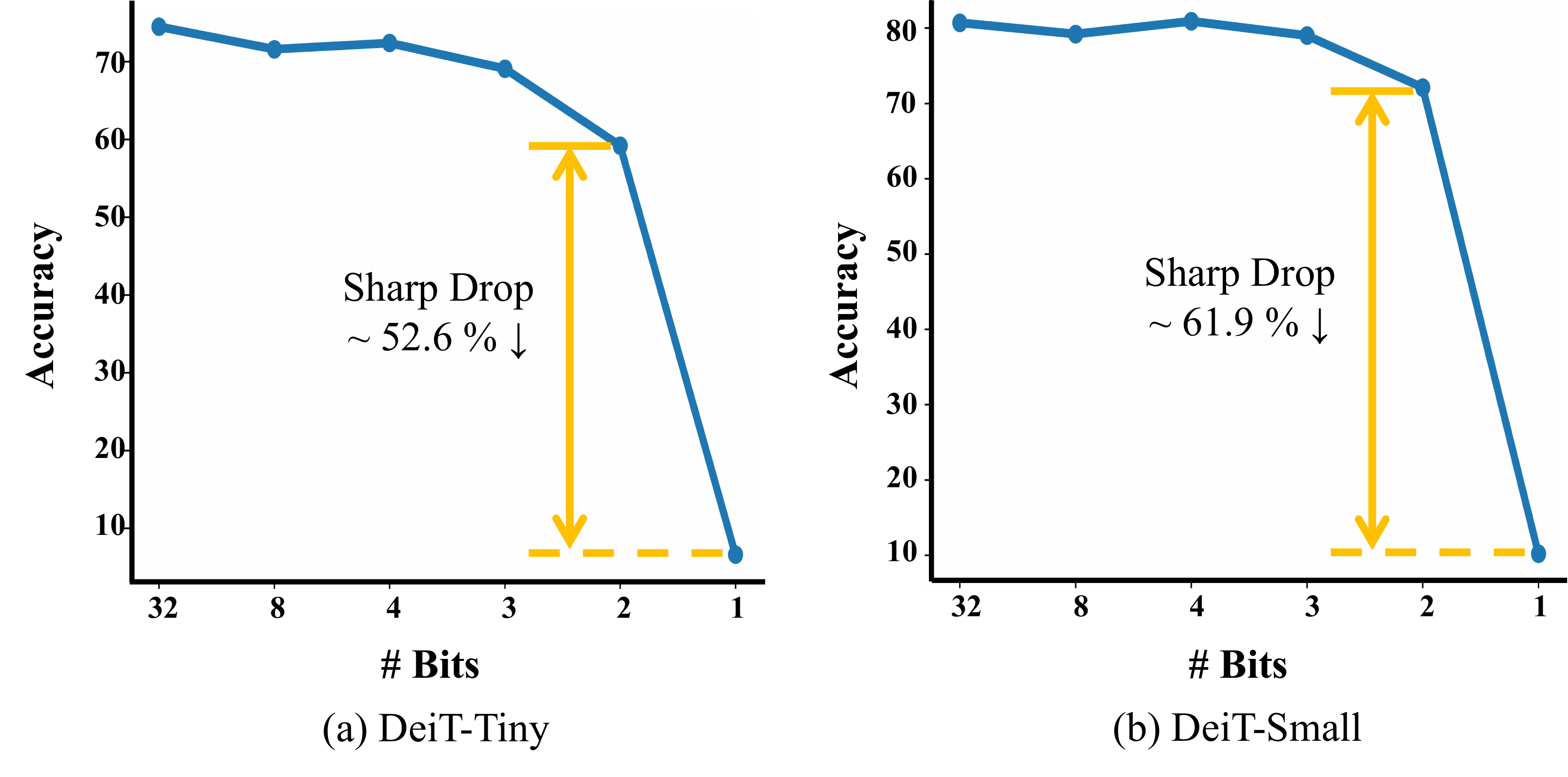}
    \caption{Performance of real-valued and quantized DeiT~\cite{touvron2021training} with varying bit-widths. We report results with  (a) DeiT-Tiny and (b) DeiT-Small on ImageNet~\cite{imagenet12}, respectively. Here 8-bit DeiT is quantized with PTQ method~\cite{lin2021fq} and 2/3/4 bit DeiT is trained with QAT method~\cite{li2022q}. The binarized ViT is conducted with the baseline method Bi-Real Net~\cite{liu2018bi}.}
    \label{fig:motivation_1}
\end{figure}

Network quantization, which represents weights and activations in a low-bit format, has got great earnestness of many researchers for its reduced memory access costs and increased compute efficiency as well as performance benefit. Using the lower-bit quantized data, in particular to the extreme 1-bit case, requires less data movement, both on-chip and off-chip, and therefore reduces memory bandwidth and saves significant energy. Existing documentary records observe 32$\times$ less network size and 58$\times$ speedups beneficial from xnor and bit-count logics for 1-bit networks~\cite{rastegari2016xnor}.
Earlier attempts~\cite{liu2021post,lin2021fq} apply post-training quantization (PTQ)~\cite{banner2019post,zhong2022fine} directly to ViTs without data-driven fine-tuning, causing sub-optimal performance, in particular to impotent 1-bit ViTs.
Therefore, by quantizing while training, quantization-aware training (QAT) methods are more congenial to 1-bit ViTs. Extensive empirical studies~\cite{liu2020reactnet,lin2020rotated,xu2022recurrent,qin2022bibert} have well demonstrated the efficacy of QAT methods in 1-bit convolutional neural networks (CNNs) or BERTs, however, the application to 1-bit ViTs remains not to be fully explored so far.

\begin{figure}[t]
    \centering
    \includegraphics[scale=.24]{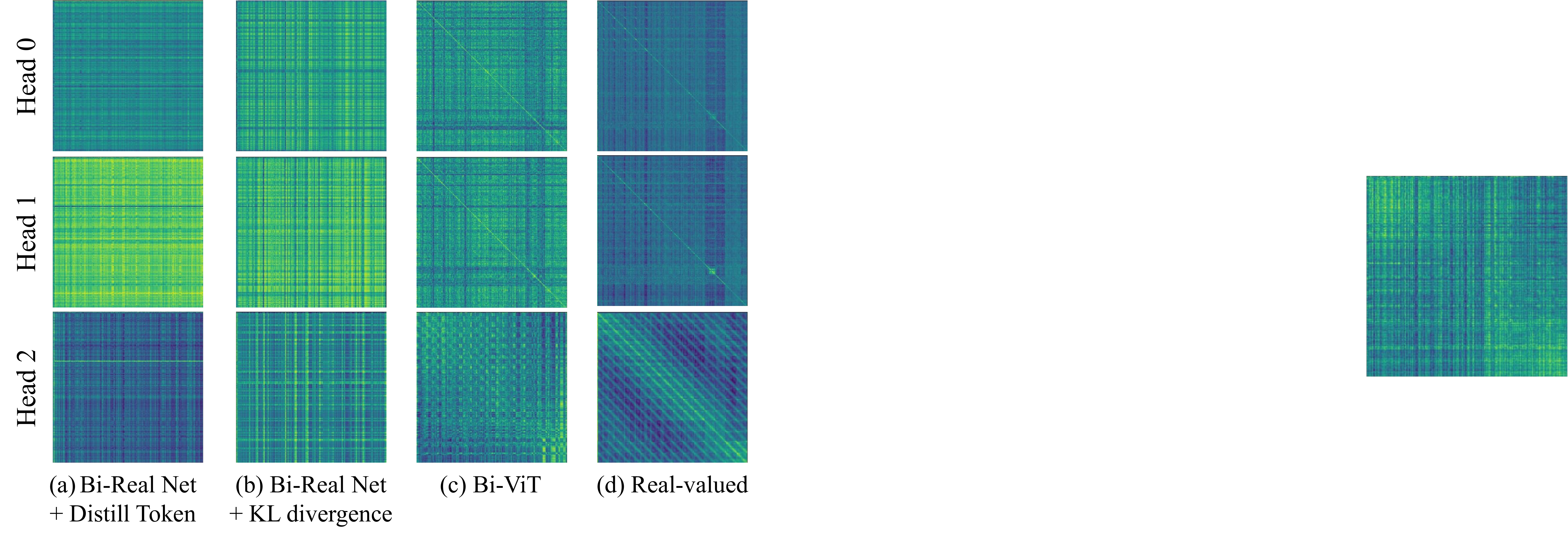}
    \caption{Visualization of the attention map before softmax in the first block of DeiT-Tiny~\cite{touvron2021training} on ImageNet~\cite{imagenet12}. From the left to right, is the baseline method~\cite{liu2018bi}, previous binarization method~\cite{xu2022recurrent},  our Bi-ViT and real-valued counterpart. }
    \label{fig:motivation_2}
\end{figure}

In this paper, we first build a fully-binarized ViT baseline, a straightforward solution constructed upon popular binarized QAT method of Bi-Real Net~\cite{liu2018bi}. Through an empirical study of this baseline, we observe significant performance drops on the ImageNet dataset~\cite{imagenet12}, as shown in Fig.\,\ref{fig:motivation_1}. For instance, extending Bi-Real Net to binarize DeiT-Tiny~\cite{touvron2021training} incurs a tremendous performance gap of 52.6\% in the Top-1 accuracy compared to the 2-bit quantized counterpart. Similar performance drops occur in DeiT-Small as well.
Delving into a deeper analysis, we find that the incompatibility of existing QAT methods mainly stems from the binarized self-attention module in ViTs, where a simple application of existing binarization methods~\cite{liu2018bi} leads to severe attention distortion, as plotted in Fig.\,\ref{fig:motivation_2}\,(a) and Fig.\,\ref{fig:motivation_2}\,(b), especially in the diagonal scores of the map which are supposed to be the most attentive.

\begin{figure*}[t]
    \centering
    \includegraphics[scale=.3]{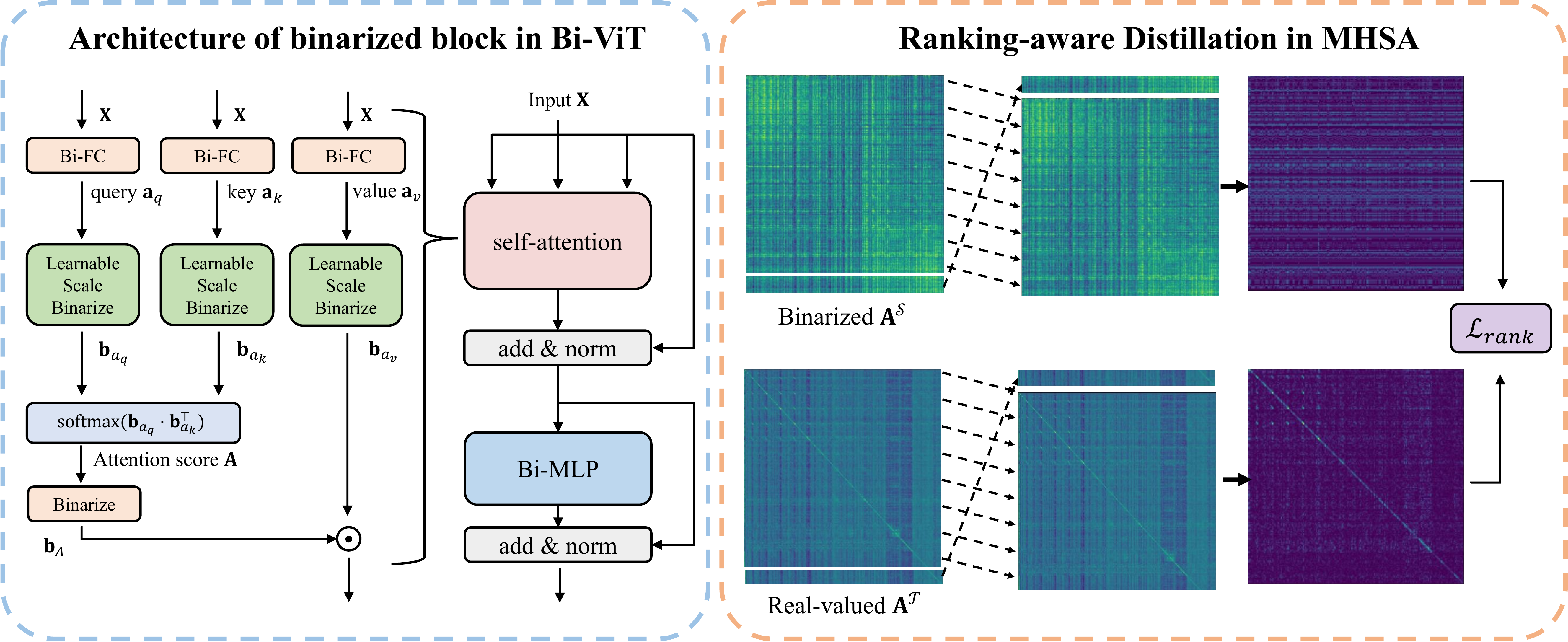}
    \caption{Overview of the proposed Bi-ViT framework. We introduce the learnable scaling factor in an architecture perspective and a ranking-aware distillation scheme incorporated in the optimization process. From left to right, we respectively show the detailed architecture of single block in Bi-ViT and the distillation framework of Bi-ViT.}
    \vspace{2mm}
    \label{fig:framework}
\end{figure*}

In this paper we dig deeper into this attention distortion problem. Through empirical analysis, we find that this phenomenon is mainly caused by gradient vanishing due to the straight-through-estimator (STE)~\cite{bengio2013estimating} and non-scaled binarization in self-attention. Meanwhile, a simple distillation utilizing distillation token in DeiT~\cite{touvron2021training} and KL-divergence in ReActNet~\cite{liu2020reactnet} is ineffective in dismissing the ranking disorder, since it neglects the relative order of the attention map between the binarized ViTs and their real-valued counterpart. 
To address the aforementioned issues, a fully-binarized ViT (Bi-ViT) is developed by reactivating the vanished gradients through a learnable scaling factor in self-attention and a ranking-aware distillation to further effectively rectify the disordered ranking of attention (see the overview in Fig.~\ref{fig:framework}). 
In addition, we also provide both empirical and theoretical analysis about how our method can rectify the distorted attention and thus promote the optimization of Bi-ViT.
%
The contributions of our work are summarized as:

\begin{itemize}
    \item We identify the bottleneck of a fully-binarized ViT through empirical analyses and formulate the problem in a theoretical perspective. Based on these, we introduce learnable head-wise scaling factor into binarized self-attention to reactivate the vanished gradients. 
    \item We develop a ranking-aware distillation scheme to eliminate attention distortion. 
    Our distillation method fully utilizes the ranking-aware knowledge  from the real-valued teacher to promote the optimization of Bi-ViT.
    \item Our Bi-ViT is   the first  promising way to push the limit of ViT quantization to the fully-binarized version. Extensive experiments on the ImageNet benchmark demonstrate that Bi-ViT surpasses both the baseline and prior binarized methods by a significant margin, achieving a remarkable acceleration rate of up to 61.5$\times$.
\end{itemize}

\section{Related Work}

\noindent\textbf{Vision Transformer}. 
Unlike traditional CNN-based models, ViTs are capable of capturing long-range visual relationships through the self-attention mechanism, and offer a more generalizable paradigm without inductive bias specific to images. The starting ViT~\cite{dosovitskiy2020image} views an image as a sequence of 16 $\times$ 16 patches and uses a unique class token to predict the classification, yielding promising results. Subsequently, many works, such as DeiT~\cite{touvron2021training} and PVT~\cite{wang2021pyramid}, have improved upon ViT, making it more efficient and applicable to downstream tasks.
However, these high-performing ViTs have also accompanied with a significant number of parameters and high computational overhead, limiting their widespread applications. Thus, designing smaller and faster ViTs has become a new trend. LeViT~\cite{graham2021levit} makes progress in faster inference through down-sampling, patch descriptors, and a redesign of the Attention-MLP block. DynamicViT~\cite{rao2021dynamicvit} proposes a dynamic token sparsification framework to progressively and dynamically prune redundant tokens, achieving a competitive complexity and accuracy trade-off. Evo-ViT~\cite{xu2021evo} proposes a slow-fast updating mechanism that ensures information flow and spatial structure, reducing both the training and inference complexity. While the aforementioned works focus on efficient model design, this paper aims to boost compression and acceleration through binarization.

\noindent\textbf{Network Binarization}. 
BinaryNet is a technique originally proposed to train convolutional neural networks (CNNs) with binary weights. BinaryConnect~\cite{courbariaux2015binaryconnect} is the precursor to BinaryNet, where the parameters are binary while the activations remain in full-precision states. Local binary convolution layers (LBC)~\cite{juefei2017local} were introduced to binarize the non-linear activations, and XNOR-Net~\cite{rastegari2016xnor} was introduced to improve convolution efficiency by binarizing the weights and inputs of convolution kernels. Bi-Real Net~\cite{liu2018bi} explores a new variant of residual structure to preserve the information of real activations before the sign function, with a tight approximation to the derivative of the non-differentiable sign function. Real-to-binary~\cite{martinez2020training} re-scales the feature maps on the channels according to the input before binarized operations and adds an SE-Net~\cite{hu2018squeeze} like gating module. ReActNet~\cite{liu2020reactnet} replaces the conventional PReLU and the sign function of the BNNs with RPReLU and RSign with a learnable threshold, thus improving the performance of BNNs. RBONN~\cite{xu2022recurrent} introduces a recurrent bilinear optimization to address the asynchronous convergence problem for BNNs, which further improves the performance of BNNs. These techniques improve the efficiency and accuracy of binary neural networks (BNNs) and allow them to be applied in practical applications. 
Majorities of these techniques consider non-scaled binarization in activations, which is beneficial to conventional CNNs while causing gradient mismatch issue for the pecularity of self-attention mechanism in ViTs.



\section{Background}
\subsection{Multi-Head Self-Attention and Binarization}
%
%
For a multi-head self-attention (MHSA) module, we denote its query, key, and value set as $\{{\bf a}_{\{q, k, v\}} \in \mathbb{R}^{h \times N \times d}\}$, where $h$ denotes head number, $N$ and $d$ represent the patch and channel numbers of each head. Specifically, $N = (W_{in} // W_{in}^P) \times (H_{in} // H_{in}^P)$ where $W_{in}$ and  $H_{in}$ are the width and height of the feature, $W_{in}^P$, $H_{in}^P$ are the width and height of patch maps respectively. Then, the attention score ${\bf A}$ and MHSA module output ${\bf a}_{out}$ are computed as follows~\cite{vaswani2017attention}:
\begin{equation}
\begin{aligned}
&{\bf A} = \operatorname{softmax}[({\bf a}_{q} \cdot {\bf a}_k^\top) / \sqrt{d}], \\
&{\bf a}_{out} = {\bf A} \cdot {\bf a}_v^\top, 
\end{aligned}
\label{float_mhsa}
\end{equation}
where $\operatorname{softmax}(\cdot)$ represents the softmax operation. Intending to represent query, key, value and attention score, {\em i.e.}, ${\bf a}_{q}$, ${\bf a}_{k}$, ${\bf a}_{v}$ and ${\bf A}$, in a 1-bit format, Eq.\,(\ref{float_mhsa}) changes into:
\begin{equation}
    \begin{aligned}
    &{\bf A} = \operatorname{softmax}[({\bf b}_{a_{q}} \cdot {\bf b}_{a_{k}}^\top) / \sqrt{d}], \\
    &{\bf a}_{out} = {\bf b}_{A} \cdot {\bf b}_{a_{v}}^\top. 
    \end{aligned}
    \label{bi_mhsa}
\end{equation}

We follow the common network binarization methods~\cite{rastegari2016xnor} that use the sign function ${\bf b}_{\cdot} = \operatorname{sign}(\cdot)$ in the binary forward pass, and STE~\cite{bengio2013estimating} $\frac{\partial {\bf b}_{\cdot}}{\partial \cdot} = 1_{|\cdot| \le 1}$ to compute the gradient for sign function in its backward pass. We omit the non-linear function here for simplicity. 
For all the projection and linear layers in binarized ViTs, we conduct binarization following~\cite{qin2022bibert,liu2018bi} as ${{\bf a}_{out}} = {\bf b}_{a_{in}} \cdot (\bm{\alpha}_{w} \circ {\bf b}_{w})^\top = \bm{\alpha}_{w} \circ ({\bf b}_{a_{in}} \cdot {\bf b}_w^\top)$ where $\bm{\alpha}_{w} = \{ {\alpha}_{w}^1, {\alpha}_{w}^2, ..., {\alpha}_{w}^{C_{out}}\} \in \mathbb{R}^{C_{{out}}}_+$ is known as the channel-wise scaling factor vector~\cite{rastegari2016xnor} and $\circ$ represents channel-wise channel-wise multiplication. 
The matrix multiplication process, {\em i.e.}, ${\bf b}_{a_{in}} \cdot {\bf b}_w^\top$, can be executed by the efficient XNOR and Bit-count instructions 
on edge devices. 

\begin{figure}[t]
    \centering
    \hspace{-4mm}
    \includegraphics[scale=.24]{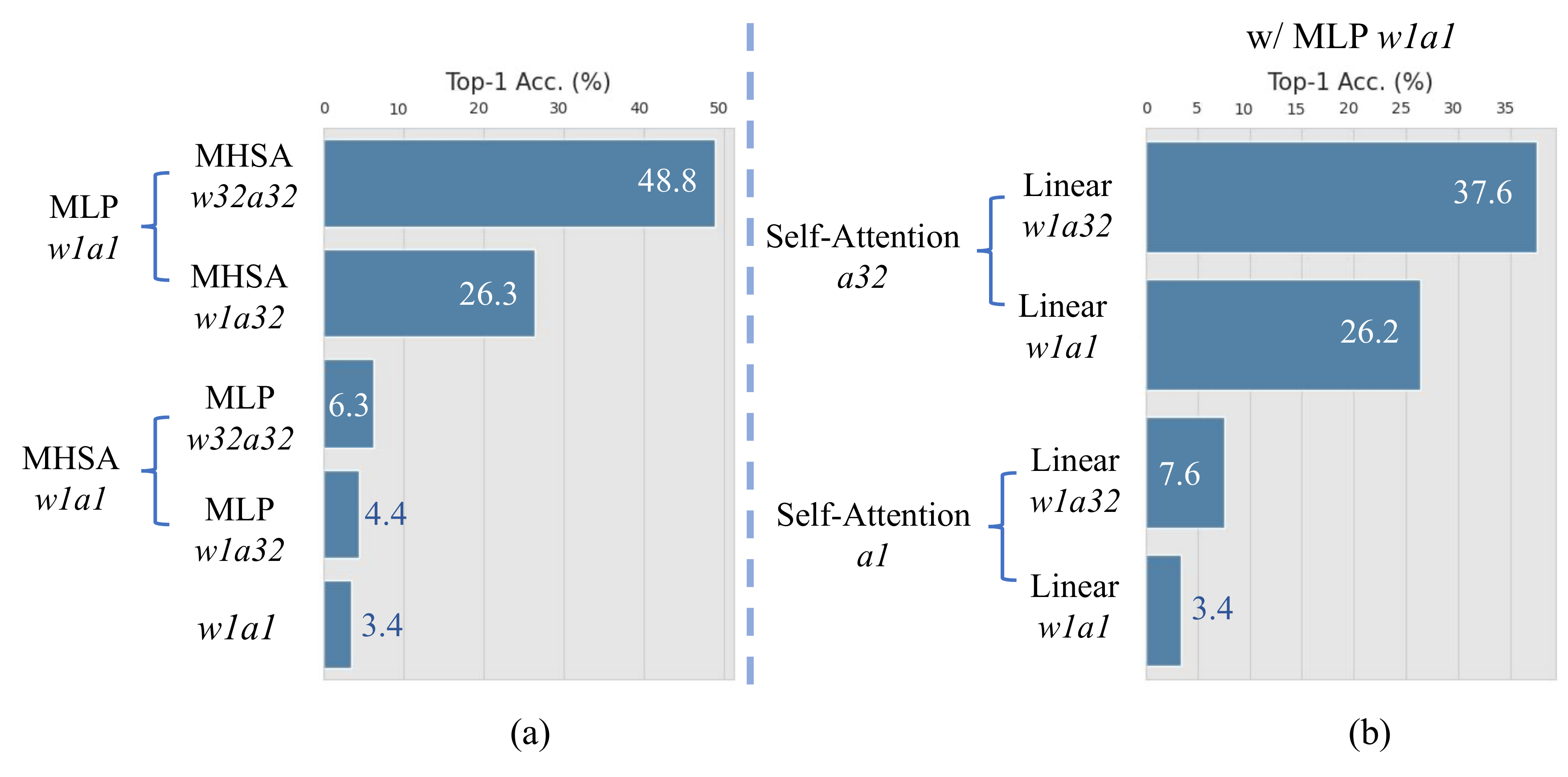}
    \caption{Performance of fully-binarized DeiT-Tiny on ImageNet~\cite{imagenet12} with different binarized/real-valued settings. }
    \label{fig:bottleneck}
\end{figure}

\subsection{Bottleneck of Fully-Binarized ViTs}
\label{sec:challenges}
%
%

%
The high-performing ViTs are built on premise of transformer's supreme ability to model the long-range relationships thanks to the attention mechanism within the MHSA module. Unfortunately, a binarized version of weights and inputs significantly weakens the representation ability. In addition, the sign function and clamp operation also damage the optimization of backpropagation.
To be more evident, we perform quantitative ablative experiments where we replace weights or activations in each module of the real-valued DeiT-Tiny~\cite{touvron2021training} with a binarized one and report the resulting Top-1 accuracy drop on the ImageNet dataset~\cite{imagenet12} after a total of 50 training epochs. Fig.\,\ref{fig:bottleneck} reports the results and we go on a deeper analysis below.

\noindent\textbf{Module Degradation}. 
By gradually replacing the multilayer perceptron (MLP) and MHSA modules with real-valued weights or activations, we have discovered that maintaining the MLP as ``{\em w1a1}'' (all weights and activations in the MLP are binarized) still results in satisfactory performance. For instance, keeping MLP as ``{\em w1a1}'' while keeping MHSA as ``{\em w1a32}'' obtains 26.3\% Top-1 accuracy, which might be acceptable comparing to the 55.2\% of real-valued DeiT-Tiny when taking into consideration 47.3$\times$ acceleration rates.
On the contrast, when maintaining MHSA module as ``{\em w1a1}'', we observe a significant drop in performance. To be more specific, even when the MLP was maintained as ``{\em w32a32}'', we still observe a significant 50.8\% decrease in Top-1 accuracy (from 55.2\% to 4.4\%). This result indicates that using binarized weights and activations in the MHSA module can have a substantial negative impact on the model's performance, even when other parts retain in real-valued states.

\noindent\textbf{Operation Degradation}. 
To better understand the impact of fully-binarized ViT's performance, we conduct further analyses by examining the operations within the MHSA module. Specifically, when we maintain the self-attention activations in Eq.\,(\ref{float_mhsa}) as real-valued (``{\em a32}''), we observe only a relatively small decrease in performance from 48.8\% to 37.6\%. However, when the self-attention activations in Eq.\,(\ref{bi_mhsa}) are binarized, significant drops in accuracy occur from 48.8\% to 7.6\%. This finding highlights the importance of the self-attention process within the MHSA module and suggests more efforts to mitigate the negative impact of binarization on the MHSA module.


\subsection{Gradient Mismatch in Self-Attention}
\label{sec:mismatch}
With conclusion from the experimental results in Sec.\,\ref{sec:challenges} that self-attention process, {\em i.e.}, Eq.\,(\ref{bi_mhsa}), is the most critical part causing the performance drops. We attempt to analyze the underlying reasons for this phenomenon from an optimization perspective. For simplicity, we derive the gradient mismatch in ${\bf a}_q$ as an example, and the analysis can be applicable to explain ${\bf a}_k$ as well. 
We first represent the features before $\operatorname{softmax}(\cdot)$ in Eq.\,(\ref{bi_mhsa}) as:
\begin{equation}
    \begin{aligned}
    {\bf p} &= ({\bf b}_{a_{q}} \cdot {\bf b}_{a_{k}}^\top.) / \sqrt{d}.  
    \end{aligned}
    \label{probability}
\end{equation}

The gradient of ${\bf a}_{q}^{h_i, n, c}$ {\em w.r.t.} ${\bf A}$ is formulated as:
\begin{equation}
\small
    \begin{aligned}
        \frac{\partial {\bf A}}{\partial {\bf a}_{q}^{h_i, n, c}} = \frac{\partial {\bf A}}{\partial {\bf p}^{h_i, n, n'}} \cdot \frac{\partial {\bf p}^{h_i, n, n'} }{\partial {\bf b}_{{a}_{q}}^{h_i, n, c} } \cdot \frac{\partial {\bf b}_{{a}_{q}}^{h_i, n, c} }{\partial {\bf a}_{q}^{h_i, n, c} },         
    \end{aligned}
    \label{gradient_q}
\end{equation}
where $h_i \in \mathbb{R}^{h}$, $n$ \& $n' \in \mathbb{R}^{N}$, $c \in \mathbb{R}^{d}$ and the gradient of ${\bf a}_{k}$ is likewise. The explicit form of the first item $\frac{\partial {\bf A}}{\partial {\bf p}^{h_i, n, n'}}$ in Eq.\,(\ref{gradient_q}) is:
\begin{equation}
    \begin{aligned}
        \frac{\partial {\bf A}}{\partial {\bf p}_{h_i, n, n'}} &= \frac{\partial \operatorname{softmax}({\bf p}_{h_i, n, n'})}{\partial {\bf p}_{h_i, n, n'}} \\
        &= {\bf A}_{h_i, n, n'} \otimes (1 - {\bf A}_{h_i, n, n'}), 
    \end{aligned}
    \label{first_item}
\end{equation}
where $\otimes$ denotes Hadamard product. 
And the second item is formulated as: 
\begin{equation}
\small
    \begin{aligned}
        \frac{\partial {\bf p}^{h_i, n, n'} }{\partial {\bf b}_{{a}_{q}}^{h_i, n, c} } &=  \frac{\partial {\bf b}_{{a}_{q}}^{h_i, n, c} \cdot {\bf b}_{{a}_{k}}^{\top{h_i, c, n'}}}{\partial {\bf b}_{{a}_{q}}^{h_i, n, c} } \\
        &= {\bf b}_{{a}_{k}}^{\top{h_i, c, n'}}, 
    \end{aligned}
    \label{second_item}
\end{equation}
result of which is therefore correlated with ${\bf b}_{{a}_{k}}$. The third item is solved through STE~\cite{bengio2013estimating} as:
\begin{equation}
    \begin{aligned}
        \frac{\partial {\bf b}_{{a}_{q}}^{h_i, n, c} }{\partial {\bf a}_{q}^{h_i, n, c} } = {\bf 1}_{|{\bf a}_{q}^{h_i, n, c}| \le 1}. 
    \end{aligned} 
    \label{third_item}
\end{equation}

Combing Eq.\,(\ref{first_item})$-$Eq.\,(\ref{third_item}), we have the final gradient form in fully-binarized ViTs as:
\begin{equation}
\small
    \begin{aligned}
        \frac{\partial {\bf A}}{\partial {\bf a}_{q}^{h_i, n, c}} &= \frac{\partial {\bf A}}{\partial {\bf p}^{h_i, n, n'}} \cdot \frac{\partial {\bf p}^{h_i, n, n'} }{\partial {\bf b}_{{a}_{q}}^{h_i, n, c} } \cdot \frac{\partial {\bf b}_{{a}_{q}}^{h_i, n, c} }{\partial {\bf a}_{q}^{h_i, n, c} } \\
        &= {{\bf A}_{h_i, n, n'} (1 - {\bf A}_{h_i, n, n'})} \cdot {\bf b}_{{a}_{k}}^{h_i, c, n'} \cdot {\bf 1}_{|{\bf a}_{q}^{h_i, n, c}| \le 1}. 
    \end{aligned}
    \label{gradient_final}
\end{equation}

\begin{figure}[!t]
    \centering
    \includegraphics[scale=.26]{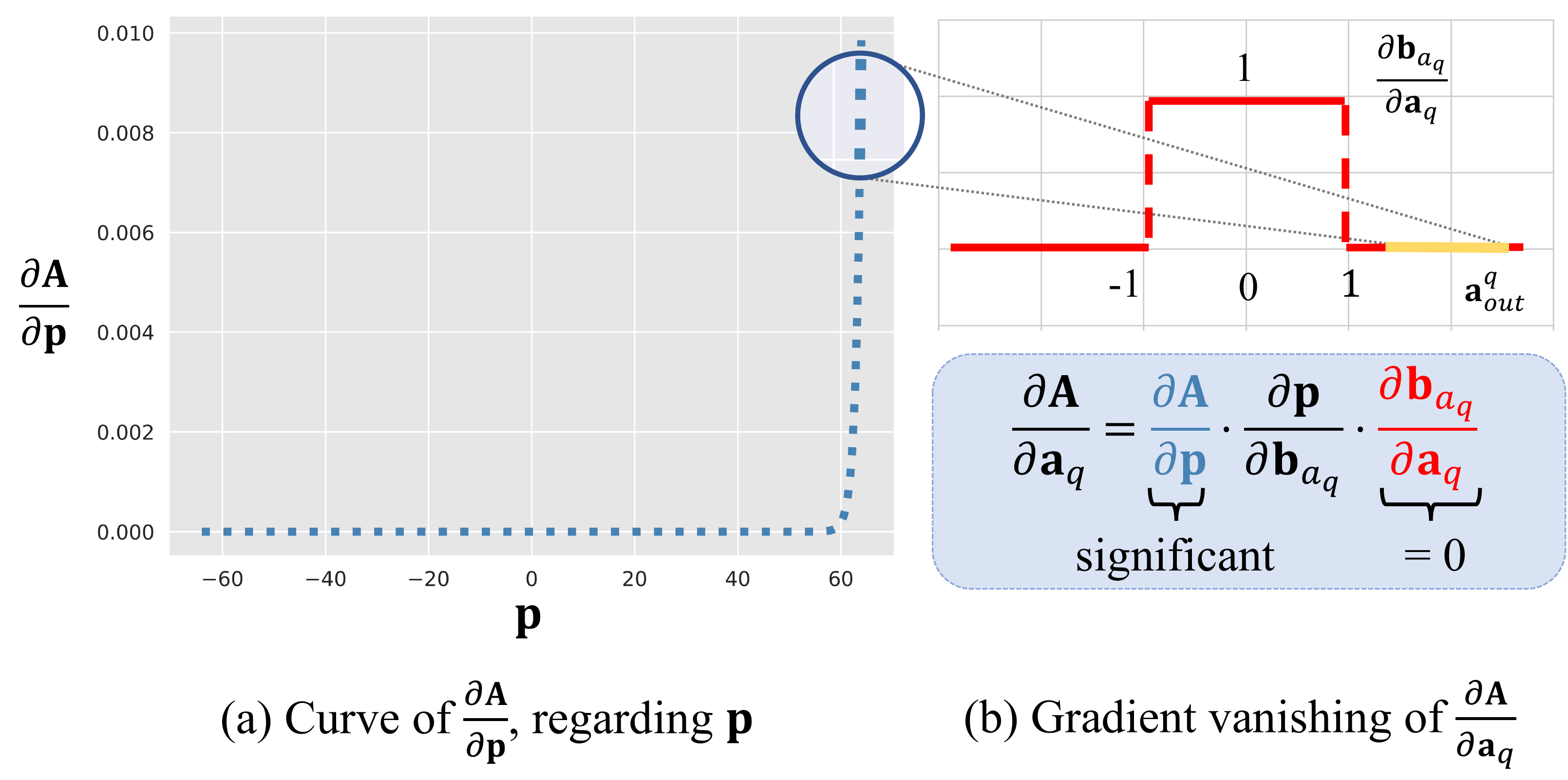}
    \caption{Illustration of gradient mismatch between Eq.\,(\ref{first_item}) and Eq.\,(\ref{third_item}).}
    \label{fig:mismatch}
\end{figure}

Considering ${\bf b}_{{a}_{q}}^{h_i, n, :}  =  [1, \cdots, 1]$ and  $\cdot {\bf b}_{{a}_{k}}^{h_i, n', :}  =  [1, \cdots, 1]$
as the extreme condition, ${\bf b}_{{a}_{q}}^{h_i, n, :} \cdot {\bf b}_{{a}_{k}}^{\top{h_i, :, n'}} = d$.  
Therefore, a specific element in ${\bf b}_{a_q} \cdot {\bf b}^\top_{a_k}$ is $\in \{- d, \cdots, d\}$. 
We plot the curve of a specific element in the first item between $[-64, 64]$ in Fig.\,\ref{fig:mismatch}\,(a) as $d = 64$ in DeiT-Tiny~\cite{touvron2021training}. We observe $\frac{\partial {\bf A}}{\partial {\bf p}_{h_i, n, n'}}$ sharply magnified when ${\bf p}_{h_i, n, n'}$ increases.
As shown in Fig.\,\ref{fig:mismatch} (b), when ${\bf p}_{h_i, n, n'}$ has a large magnitude, $|{\bf a}_{q}| > 1$ and $\frac{\partial {\bf b}_{{a}_{q}}^{h_I, n, c} }{\partial {\bf a}_{q}^{h_i, n, c} } = 0$ . Thus the multiplication of these two items leads to $\frac{\partial {\bf A}}{\partial {\bf a}_{q}^{h_i, n, c}} = 0$, likewise for ${\bf a}_{k}$.
Therefore we formulate the gradient mismatch phenomenon in the aforementioned theoretical analysis. And such gradient mismatch leads to distorted gradient in the optimization of ${\bf a}_{q}$ \& ${\bf a}_{k}$ and therefore degrades performance of fully-binarized ViTs.

\section{Our Bi-ViT}
In this section, we propose to dismiss the affect of gradient mismatch mentioned in Sec.\,\ref{sec:mismatch} from perspectives of gradient approximation in Sec.\,\ref{sec:learnable_scale} and intermediate distillation in Sec.\,\ref{sec:distillation}.

\subsection{Learnable Head-wise Scaling Factor}
\label{sec:learnable_scale}
As one of the solution to the above mentioned problem, we propose a head-wise scaling factor binarization scheme for the self-attention process, where the scaling factors are learned during training to first modify the gradient clip range in Fig.\,\ref{fig:mismatch}(b). Eq.\,(\ref{bi_mhsa}) is changed into:
\begin{equation}
    \begin{aligned}
    \tilde{\bf A} &= \operatorname{softmax}(\tilde{\bf p}), \\
    \tilde{\bf p} &= (\bm{\alpha}_{q} \otimes \bm{\alpha}_{k}) \circ ({\bf b}_{{a}_{q}} \cdot {\bf b}_{{a}_{k}}^\top) / \sqrt{d} \\
    &= \bm{\alpha}_{q;k} \circ ({\bf b}_{{a}_{q}} \cdot {\bf b}_{{a}_{k}}^\top) / \sqrt{d}, 
    \end{aligned}
    \label{bivit_attention}
\end{equation}
and 
\begin{equation}
\small
    \begin{aligned}
    \tilde{\bf a}_{out} &= (\bm{\alpha}_{A} \circ {\bf b}_{A}) \cdot (\bm{\alpha}_{v} \circ {\bf b}_{{a}_v})^\top \\
    &= (\bm{\alpha}_{A} \otimes \bm{\alpha}_{v}) \circ ({\bf b}_{A} \cdot {\bf b}_{{a}_v}^\top) \\
    &= \bm{\alpha}_{A;v} \circ ({\bf b}_{A} \cdot {\bf b}_{{a}_v}^\top),
    \end{aligned}
    \label{bivit_mhsa}
\end{equation}
where ${\bf b}_{{a}_{\cdot}} = \operatorname{sign}(\frac{{\bf a}_{\cdot}}{\bm{\alpha}_{\cdot}})$, $\bm{\alpha}_{q}$, $\bm{\alpha}_{k}$, $\bm{\alpha}_{v}$ and $\bm{\alpha}_{A}$
 are the head-wise learnable scaling factors in binarized MHSA, where $\bm{\alpha}_{\{q, k, v, A\}} = \{\alpha_{\{q, k, v, A\}}^1, \alpha_{\{q, k, v, A\}}^2, \cdots, \alpha_{\{q, k, v, A\}}^h\} \in \mathbb{R}^h_+$. The second rows in Eq.\,(\ref{bivit_attention}) \& Eq.\,(\ref{bivit_mhsa}) are established since the scaling factors are aligned with the head dimension, which is independent with the matrix multiplication operation. Thus, $\bm{\alpha}_{q;k} = \{\alpha_{q;k}^1, \alpha_{q;k}^2, \cdots, \alpha_{q;k}^h\} \in \mathbb{R}^h_+$ and $\bm{\alpha}_{A;v} = \{\alpha_{A;v}^1, \alpha_{A;v}^2, \cdots, \alpha_{A;v}^h\} \in \mathbb{R}^h_+$.

 Consequently, the gradient $\frac{\partial \tilde{\bf A}}{\partial {\bf a}_{q}^{:, n, c}}$ in Eq.\,(\ref{gradient_final}) is further formulated in our Bi-ViT as:
 \begin{equation}
 \small
    \begin{aligned}
        \frac{\partial \tilde{\bf A}}{\partial {\bf a}_{q}^{h_i, n, c}} 
        &= \underbrace{{\tilde{\bf A}^{h_i, n, n'} (1 - \tilde{\bf A}^{h_i, n, n'})}}_{\frac{\partial \tilde{\bf A}}{\partial {\bf p}^{h_i, n, n'}}} \cdot \underbrace{\bm{\alpha}_{q;k}^{h_i} \circ {\bf b}_{a_{k}}^{h_i, c, n'}}_{\frac{\partial {\bf p}^{h_i, n, n'} }{\partial {\bf b}_{{a}_{q}}^{h_i, n, c} }} \cdot \underbrace{{\bf 1}_{|{\bf a}_{q}^{h_i, n, c}| \le \bm{\alpha}_{q}}}_{\frac{\partial {\bf b}_{{a}_{q}}^{h_i, n, c} }{\partial {\bf a}_{q}^{h_i, n, c} }}. 
    \end{aligned}
    \label{gradient_final_bivit}
\end{equation}

Since $\operatorname{softmax}(.)$ and $\circ$ are aligned with different dimensions, the value of Eq.\,(\ref{first_item}) remains unchanged ($\operatorname{softmax}({\bf p}) = \operatorname{softmax}(\bm{\alpha}_{q;k} \circ {\bf p})$). 
As can be seen, the threshold of gradient clip in Eq.\,(\ref{third_item}) changes from $1$ into $\bm{\alpha}_{q}$, which means that we can surpass the occurance of gradient mismatch by modifying the value of $\bm{\alpha}_{q}$. 
Note that the scaling factor ($\bm{\alpha}_{q}$) is to imitate the magnitude of the latent activations. When $\tilde{\bf p}$ has a large magnitude, {\em i.e.}, in the circled part of Fig.\,\ref{fig:mismatch} (a), $\bm{\alpha}_{q}$ also tends to be larger and ${\bf a}_{q}^{h_i, n, c}$ locates in the field that $\frac{\partial {\bf b}_{{a}_{q}}^{h_i, n, c} }{\partial {\bf a}_{q}^{h_i, n, c} } > 0$. 
Thus the vanishing gradients are  reactivated through the introduced learnable scaling factor.

\subsection{Ranking-aware Distillation for Bi-ViT}
\label{sec:distillation}
Fig.\,\ref{fig:motivation_2} illustrates a significant difference in the attention map's relative order between Bi-RealNet (a) and its real-valued counterpart (c). This difference could result in a notable decrease in performance. To address this issue during binarized training, a ranking-aware distillation in a teacher-student framework is introduced: 
\begin{equation}
    \begin{aligned}
        \mathcal{L}_{ranking} = \sum_{l=1}^{L} \| \psi({\bf A}^{\mathcal{T}}) - \psi({\bf A}^{\mathcal{S}})\|_2, 
    \end{aligned}
    \label{eq:rank_loss}
\end{equation}
where ${\bf A}^{\mathcal{T}}$ and ${\bf A}^{\mathcal{S}}$ represents the attention scores from the real-valued teacher and binarized student. $\psi(\cdot)$ denotes the function for obtaining the ranking, {\em i.e.}, relative order of an attention score, which is formulated as:
\begin{equation}
    \begin{aligned}
        \psi({\bf A}^{:, n, :}) = 
        \left\{\begin{aligned}
			{\bf A}^{:, n, :} - {\bf A}^{:, n - 1, :},    &\;\;\;\; \operatorname{if} \;\;  0 < n \le N - 1\\
			{\bf A}^{:, 0, :} - {\bf A}^{:, N - 1, :},    &\;\;\;\; \operatorname{otherwise}. 
        \end{aligned} \right. 
    \end{aligned}
    \label{eq:order}
\end{equation}
Detailed relative order computation can be seen in the right part of Fig.\,\ref{fig:framework}. We implement our Bi-ViT under the teacher-student framework~\cite{touvron2021training}, thus the final objective of our method is formulated as:
\begin{equation}
    \begin{aligned}
        \mathcal{L} = \mathcal{L}_{dist} + \lambda \mathcal{L}_{ranking}, 
    \end{aligned}
    \label{eq:final}
\end{equation}
where $\lambda$ is a hyper-parameter to balance these two loss functions.

\section{Experiments}
\label{sec:exp}
In this section, we evaluate the performance of the proposed Bi-ViT model for image classification task using popular DeiT~\cite{touvron2021training} \& Swin~\cite{liu2021swin} backbones and object detection task using Mask R-CNN~\cite{he2017mask} \& Cascade~\cite{cai2018cascade} Mask R-CNN with Swin-Tiny~\cite{liu2021swin} backbone. To the best of our knowledge, there is no publicly available source codebase on fully-binarized ViTs at this point, so we implement the baseline {\em i.e.}, Bi-Real Net~\cite{liu2018bi} methods by ourselves. 

\subsection{Datasets and Implementation Details}
\noindent{\bf Datasets}. The experiments are carried out on the ImageNet ILSVRC12 dataset~\cite{imagenet12} for image classification task and COCO dataset~\cite{coco2014} for object detection task. The ImageNet dataset is challenging due to its large scale and greater diversity. There are 1000 classes and 1.2 million training images, and 50k validation images in it. In our experiments, we use the classic data augmentation method described in~\cite{touvron2021training}.

The COCO~\cite{coco2014} dataset includes images from 80 different categories. All of our COCO dataset experiments are performed on the object detection track of the COCO {\tt trainval35k} training dataset, which consists of 80k images from the COCO {\tt train2014} dataset and 35k images sampled from the COCO {\tt val2014} dataset. We report the average precision (AP) for IoUs$\in$ [0.5:0.05:0.95], designated as mAP@[.5,.95], using COCO's standard evaluation metric. For further analyzing our method, we also report AP$_{\rm 50}$, AP$_{\rm 75}$, AP$_s$, AP$_m$, and AP$_l$.

\noindent{\bf Experimental settings}. 
In our experiments, we initialize the weights of binarized model with the corresponding pretrained real-valued model. The binarized model is trained for 300 epochs with batch-size 512 and the base learning rate $5e\!-\!4$. We do not use warm-up scheme. For all the experiments, we apply LAMB~\cite{you2019large} optimizer with weight decay set as 0, following DeiT III~\cite{touvron2022deit}. Note that we keep the patch embedding (first) layer and the classification (last) layer as real-valued, following~\cite{esser2019learned}. 

\noindent{\bf Backbone}. We evaluate our binarized method on two popular vision transformer networks: DeiT~\cite{touvron2021training} and Swin Transformer~\cite{liu2021swin}. The DeiT-Tiny, DeiT-Small, DeiT-Base, Swin-Tiny and Swin-Small are adopted as the backbone models, whose Top-1 accuracy on ImageNet dataset are 72.2\%, 79.9\%, 81.8\%, 81.2\%, and 83.2\% respectively. For a fair comparison, we utilize the official implementation of DeiT and Swin Transformer. 

\subsection{Ablation Study}
\begin{figure}[t]
\centering
\includegraphics[scale=.42]{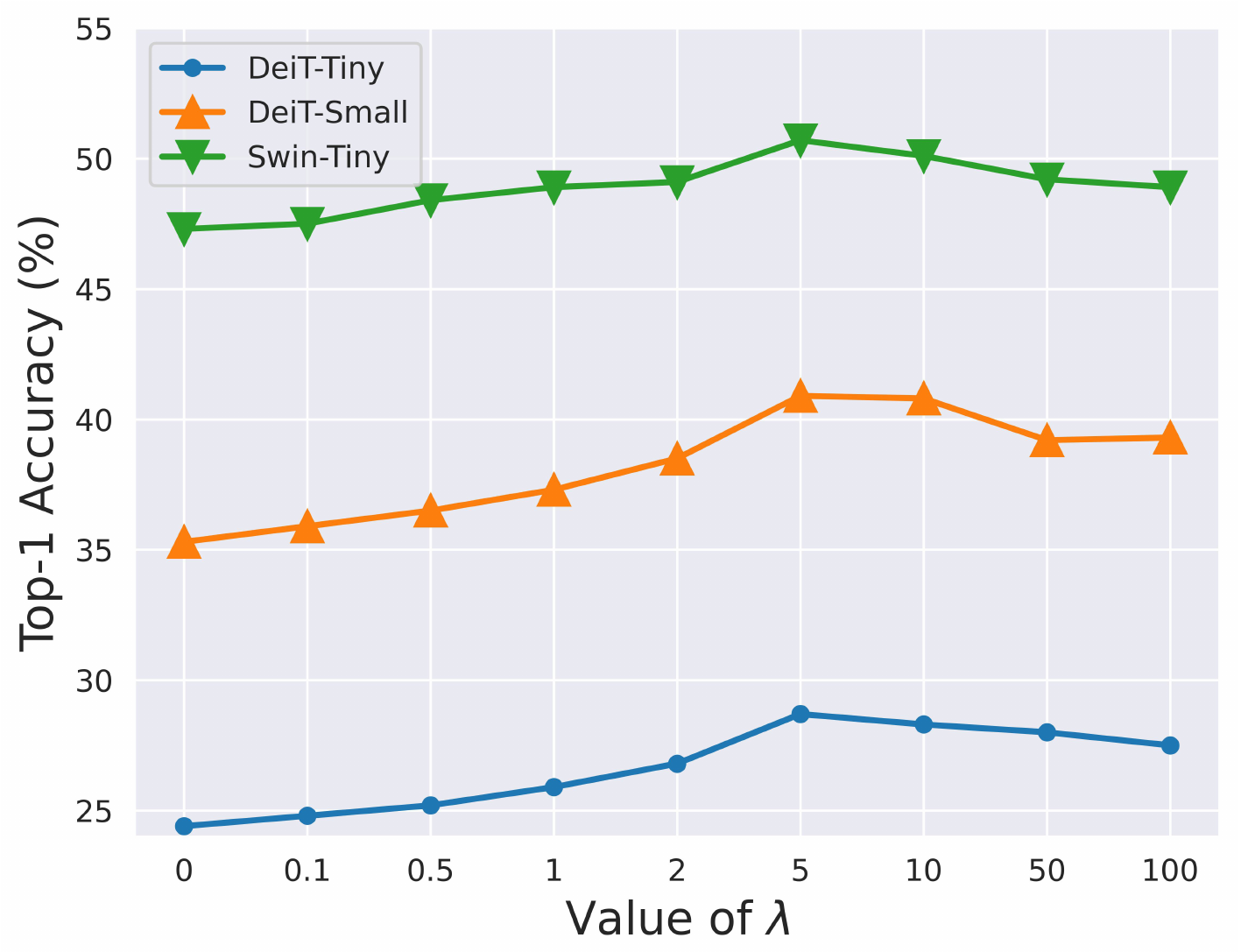}
\caption{Effect of hyper-parameter $\lambda$ on ImageNet~\cite{imagenet12}.}
\label{fig:lambda}
\end{figure}

\noindent\textbf{Hyper-parameter Selection}. 
We  $\lambda$ of Eq.\,(\ref{eq:final}) in this part, with experiments conducted on ImageNet~\cite{imagenet12} dataset. We show the model performance (Top-1 accuracy) with different setups of hyper-parameter $\lambda$ in Fig.\,\ref{fig:lambda}, in which the performances increase first and then decrease with the uplift of $\lambda$ from left to right. 
Since $\lambda$ controls the importance of $\mathcal{L}_{ranking}$, we show that the vanilla baseline ($\lambda=0$) performs worse than any versions with Ranking-aware Distillation loss
($\lambda > 0$), showing the proposed distillation scheme is necessary. With the varying value of $\lambda$, we find $\lambda=10$ boost the performance of our Bi-ViT, achieving 28.7\%, 40.9\% and 50.7\% Top-1 accuracy on ImageNet~\cite{imagenet12} with DeiT-Tiny, DeiT-Small and Swin-Tiny backbone, respectively. 


\noindent\textbf{Effectiveness of components}. 
We conduct the ablative experiments regarding the proposed components on DeiT-Tiny network. Firstly, we compose the baseline network using the binarization method following Bi-Real Net~\cite{liu2018bi}. As shown in the third row of Tab.\;\ref{tab:ablation}, the baseline networks only obtains 6.6\% Top-1 accuracy, which is far from satisfactory. With the introduction of our first novelty, {\em i.e.}, learnable scaling factor (LSF), the baseline network is boosted by 17.8\%, achieving 24.4\% Top-1 accuracy. We also observe the other contribution Ranking-aware Disitllation (RD) singly promotes the baseline network by 5.9\%, which is also significant on ImageNet dataset. By combining the two main contributions together, we get Bi-ViT, outperforming the vanilla baseline by 22.1\%.


\begin{table}[h]
\centering
\caption{Evaluating the components of Bi-ViT based on DeiT-Tiny~\cite{touvron2021training} backbone. ``\#Bits'' denotes the bit-width of weights and activations. We report the  Top-1$_{\rm (\%)}$ accuracy performances. }
\vspace{2mm}
\renewcommand\arraystretch{1.3}
\begin{tabular}{lll}
\toprule
Method                                           & \#Bits  & Top-1$_{\rm (\%)}$     \\ \hline
Real-valued                                      & 32-32         & 72.1           \\ \hline
Baseline (Bi-Real Net~\cite{liu2018bi})                                        & 1-1           & 6.6            \\
+ Learnable Scaling Factor (LSF)                 & 1-1           & 24.4$^{\color{blue} + {\rm 17.8}}$           \\
+ Ranking-aware Distillation (RD)                & 1-1           & 12.5$^{\color{blue} + {\rm 5.9}}$           \\ \cdashline{1-3}
\textbf{+ LSF + RD (Bi-ViT)}                     & 1-1           & \textbf{28.7}$^{\color{blue} + {\bf 22.1}}$  \\ \bottomrule
\end{tabular}
\label{tab:ablation}
\end{table}

\begin{table*}[t]
\centering
\caption{Experiments with DeiT~\cite{touvron2021training} and Swin~\cite{liu2021swin} on ImageNet~\cite{imagenet12}. ``\#Bits'' denotes the bit-width of weights and activations. We report the  Top-1$_{\rm (\%)}$ and  Top-5$_{\rm (\%)}$ accuracy performances. The \textbf{bold} denotes the best result with binarized weights and activations.}
\setlength{\tabcolsep}{3mm}{\begin{tabular}{lllllll}
\toprule
Network                  & Method                                 & \#Bits          & Size$_{\rm (MB)}$      & OPs$_{\rm (10^8)}$        & Top-1$_{(\%)}$ & Top-5$_{(\%)}$ \\ \hline
                         & Real-valued                            & 32-32                  & 22.8                   & 12.3                   & 72.2	     & 91.1       \\ \cline{2-7}
                         &                                        & 4-4                    & 3.0                    & 1.6                    & 74.3       & 91.7       \\
                         &                                        & 3-3                    & 2.3                    & 0.8                    & 71.5       & 91.2       \\
                         & \multirow{-3}{*}{Q-ViT~\cite{li2022q}} & 2-2                    & 1.7                    & 0.4                    & 59.0       & 81.8       \\ \cdashline{2-7}
                         & BiBERT~\cite{qin2022bibert}            &                        & \multirow{4}{*}{1.0}   & \multirow{4}{*}{0.2}   & 5.9        & 16.0         \\
                         & RBONN~\cite{xu2022recurrent}           &                        &                        &                        & 6.3      & 16.9         \\
                         & Bi-Real Net~\cite{liu2018bi}           &                        &                        &                        & 6.6      & 17.1         \\
\multirow{-8}{*}{DeiT-Tiny} & \textbf{Bi-ViT}                     & \multirow{-4}{*}{1-1}  &                        &                        & \textbf{28.7}$^{\color{blue} + {\bf 22.1}}$  & \textbf{51.7}$^{\color{blue} + {\bf 34.6}}$           \\ \hline
                         & Real-valued                            & 32-32                  & 88.2                   & 45.5                   & 79.9       & 95.0       \\ \cline{2-7}
                         &                                        & 4-4                    & 11.4                   & 5.8                    & 80.9       & 94.9       \\
                         &                                        & 3-3                    & 8.7                    & 3.0                    & 79.0       & 94.2       \\
                         & \multirow{-3}{*}{Q-ViT~\cite{li2022q}} & 2-2                    & 6.0                    & 1.5                    & 72.1       & 90.3       \\ \cdashline{2-7}
                         & BiBERT~\cite{qin2022bibert}            &                        & \multirow{4}{*}{3.4}   & \multirow{4}{*}{0.8}   & 17.4       & 29.7       \\
                         & RBONN~\cite{xu2022recurrent}           &                        &                        &                        & 18.5       & 30.0       \\
                         &  Bi-Real Net~\cite{liu2018bi}          &                        &                        &                        & 19.2       & 30.3       \\
\multirow{-8}{*}{DeiT-Small} & \textbf{Bi-ViT}                    & \multirow{-4}{*}{1-1}  &                        &                        & \textbf{40.9}$^{\color{blue} + {\bf 21.7}}$  & \textbf{65.0}$^{\color{blue} + {\bf 34.7}}$            \\ \hline
                         & Real-valued                            & 32-32                  & 346.2                  & 174.7                  & 81.8       & 95.6       \\ \cline{2-7}
                         &                                        & 4-4                    & 44.1                   & 22.0                   & 83.0       & 96.1       \\
                         &                                        & 3-3                    & 33.4                   & 11.1                   & 81.0       & 95.1       \\
                         & \multirow{-3}{*}{Q-ViT~\cite{li2022q}} & 2-2                    & 22.7                   & 5.7                    & 74.2       & 92.2       \\ \cdashline{2-7}
                         & BiBERT~\cite{qin2022bibert}            &                        & \multirow{4}{*}{12.1}  & \multirow{4}{*}{2.9}   & 24.5       & 36.3       \\
                         & RBONN~\cite{xu2022recurrent}           &                        &                        &                        & 26.1       & 38.6       \\
                         & Bi-Real Net~\cite{liu2018bi}           &                        &                        &                        & 26.5       & 38.8       \\
\multirow{-8}{*}{DeiT-Base} & \textbf{Bi-ViT}                     & \multirow{-4}{*}{1-1}  &                        &                        & \textbf{47.3}$^{\color{blue} + {\bf 20.8}}$  & \textbf{72.8}$^{\color{blue} + {\bf 34.0}}$            \\ \hline
                         & Real-valued                            & 32-32                  & 114.2                  & 44.9                   & 81.2       & 95.5       \\ \cline{2-7}
                         &                                        & 4-4                    & 14.6                   & 5.8                    & 82.5       & 97.3       \\
                         &                                        & 3-3                    & 11.2                   & 3.0                    & 80.9       & 96.1       \\
                         & \multirow{-3}{*}{Q-ViT~\cite{li2022q}} & 2-2                    & 10.0                   & 1.6                    & 74.7       & 92.5       \\ \cdashline{2-7}
                         & BiBERT~\cite{qin2022bibert}            &                        & \multirow{4}{*}{4.2}   & \multirow{4}{*}{0.8}   & 34.0       & 46.9       \\
                         & RBONN~\cite{xu2022recurrent}           &                        &                        &                        & 33.8       & 46.7       \\
                         & Bi-Real Net~\cite{liu2018bi}           &                        &                        &                        & 34.1       & 46.9       \\
\multirow{-8}{*}{Swin-Tiny} & \textbf{Bi-ViT}                     & \multirow{-4}{*}{1-1}  &                        &                        & \textbf{55.5}$^{\color{blue} + {\bf 21.4}}$   & \textbf{79.4}$^{\color{blue} + {\bf 32.5}}$             \\ \hline
                         & Real-valued                            & 32-32                  & 199.8                  & 87.5                   & 83.2       & 96.2       \\ \cline{2-7}
                         &                                        & 4-4                    & 25.3                   & 11.1                   & 84.4       & 98.3       \\
                         &                                        & 3-3                    & 19.2                   & 5.6                    & 82.7       & 97.5       \\
                         & \multirow{-3}{*}{Q-ViT~\cite{li2022q}} & 2-2                    & 13.0                   & 2.9                    & 76.9       & 94.9       \\ \cdashline{2-7}
                         & BiBERT~\cite{qin2022bibert}            &                        & \multirow{4}{*}{6.9}   & \multirow{4}{*}{1.5}   & 39.4       & 53.0       \\
                         & RBONN~\cite{xu2022recurrent}           &                        &                        &                        & 39.0       & 52.7       \\
                         & Bi-Real Net~\cite{liu2018bi}           &                        &                        &                        & 39.2       & 52.8       \\
\multirow{-8}{*}{Swin-Small} & \textbf{Bi-ViT}                    & \multirow{-4}{*}{1-1}  &                        &                        & \textbf{60.7}$^{\color{blue} + {\bf 21.5}}$   & \textbf{83.9}$^{\color{blue} + {\bf 31.1}}$             \\ \bottomrule
\end{tabular}}
\label{tab:imagenet}
\end{table*}

\subsection{Results on Image Classification}
The experimental results are shown in Tab.~\ref{tab:imagenet}. 
We compare our method with 1-bit methods including BiBERT~\cite{qin2022bibert}, RBONN~\cite{xu2022recurrent}, and Bi-Real Net~\cite{liu2018bi}  based on the same frameworks for the task of image classification with the ImageNet dataset. 
We also report the classification performance of the low-bit training-aware quantization method Q-ViT~\cite{li2022q} for further reference. We use model size and OPs following \cite{liu2018bi} in comparison to other bit-widthe models for further reference.
We firstly evaluate the proposed method on DeiT models.

\begin{table*}[t]
\centering
\caption{Experiments with Mask R-CNN~\cite{he2017mask} and Cascade R-CNN~\cite{cai2018cascade} using Swin~\cite{liu2021swin} backbones on COCO~\cite{coco2014}. ``\#Bits'' denotes the bit-width of weights and activations. We report the AP (\%) with different IoU threshold and AP for objects in various sizes. The \textbf{bold} denotes the best result with binarized weights and activations.}
\vspace{2mm}
\renewcommand\arraystretch{1.1}
\begin{tabular}{lllllllllll}
\toprule
Framework                                       & Backbone                        & Method      & \# Bits   & Size$_{\rm (MB)}$  & AP  & AP$_{50}$ & AP$_{75}$ & AP$_s$ & AP$_m$ & AP$_l$ \\ \hline
\multirow{8}{*}{Mask R-CNN~\cite{he2017mask}}   
& \multirow{8}{*}{Swin-Tiny} & Real-valued & 32-32       & 191.3                  & 43.7          & 66.6          & 47.7          & 28.5          & 47.0          & 57.3          \\
\cline{3-11}
& &                              & 4-4                   & 94.9                   & 43.3          & 66.3          & 47.1          & 28.2          & 46.5          & 57.5          \\
& &                              & 3-3                   & 91.4                   & 40.1          & 63.5          & 43.9          & 25.4          & 42.4          & 54.9          \\
& & \multirow{-3}{*}{Q-ViT~\cite{li2022q}}  & 2-2        & 88.0                   & 30.2          & 53.7          & 33.4          & 15.2          & 32.0          & 45.2          \\ \cdashline{3-11}
& & BiBERT~\cite{qin2022bibert}  & \multirow{4}{*}{1-1}  & \multirow{4}{*}{84.5}  & 8.9           & 25.0          & 8.6           & 1.7           & 9.0           &  15.9         \\
& & RBONN~\cite{xu2022recurrent} &                       &                        & 9.5           & 25.2          & 8.9           & 1.9           & 9.1           &  16.0         \\
& & Bi-Real Net~\cite{liu2018bi} &                       &                        & 9.9           & 25.2          & 9.2           & 2.1           & 9.1           &  16.4         \\
& & \textbf{Bi-ViT}              &                       &                        & \textbf{20.7} & \textbf{38.7} & \textbf{19.9} & \textbf{12.0} & \textbf{20.9} & \textbf{27.6} \\ \hline
\multirow{8}{*}{\begin{tabular}[l]{@{}l@{}}Cascade \\ Mask R-CNN~\cite{cai2018cascade}\end{tabular}} 
& \multirow{8}{*}{Swin-Tiny} & Real-valued & 32-32       & 342.7                  & 48.1          & 67.1          & 52.2          & 30.4          & 51.5          & 63.1          \\ 
\cline{3-11}
& &                              & 4-4                   & 246.2                  & 47.2          & 66.4          & 52.0          & 30.0          & 51.1          & 62.8          \\ 
& &                              & 3-3                   & 242.8                  & 44.5          & 63.1          & 49.3          & 27.8          & 48.2          & 59.9          \\ 
& & \multirow{-3}{*}{Q-ViT~\cite{li2022q}}  & 2-2        & 239.3                  & 34.0          & 33.8          & 39.6          & 17.2          & 37.6          & 49.1          \\  \cdashline{3-11}
& & BiBERT~\cite{qin2022bibert}  & \multirow{4}{*}{1-1}  & \multirow{4}{*}{235.9} & 16.7          & 32.2          & 18.1          & 11.0          & 18.7          & 25.1          \\
& & RBONN~\cite{xu2022recurrent} &                       &                        & 17.5          & 32.6          & 19.0          & 11.2          & 19.0          & 25.7          \\
& & Bi-Real Net~\cite{liu2018bi} &                       &                        & 18.1          & 33.3          & 19.1          & 12.0          & 19.2          & 25.9          \\
& & \textbf{Bi-ViT}              &                       &                        & \textbf{29.0} & \textbf{45.0} & \textbf{31.1} & \textbf{16.8} & \textbf{29.4} & \textbf{39.8} \\ \bottomrule
\end{tabular}
\label{tab:coco}
\end{table*}

For DeiT-Tiny backbone, compared with other binary methods, our Bi-ViT achieves significant performance improvements. For example, our Bi-ViT surpasses the baseline Bi-Real Net~\cite{liu2018bi} by 22.1\% Top-1 accuracy, which is significant and meaningful for real-world applications. 
And it is worth noting that the proposed 1-bit model significantly compresses the DeiT-Tiny by $61.5\times$ on OPs. 
The proposed method also boosts the performance of baseline by 21.7\% with the same architecture and bit-width using DeiT-Small bacobone, a significant improvement on the ImageNet dataset. 
For larger DeiT-B, as shown in Tab.~\ref{tab:imagenet}, the performance of the proposed method outperforms the Bi-Real Net by 20.8\%, a large margin. 
Also note that the proposed 1-bit model significantly compresses the DeiT-B by 60.2$\times$ on OPs and 28.6$\times$ on model size.

Also, our method obtains convincing results on Swin-transformers. 
As shown in Tab.~\ref{tab:imagenet}, the performance of the proposed method with Swin-Tiny outperforms the baseline method by 21.4\%, a large margin. 
%
%
%
For larger Swin-Small, the performance of the proposed method outperforms the 1-bit baseline by 21.5\%. Also note that our method theoritically accelerates the network by 58.3$\times$, which demonstrates the effectiveness and efficiency of our Bi-ViT.

\subsection{Results on Object Detection}
We conduct experiments on object detection using the COCO dataset and compare our Bi-ViT with previous binary neural networks such as BiBERT \cite{qin2022bibert}, RBONN~\cite{xu2022recurrent}, and Bi-Real Net \cite{liu2018bi}. To provide a basis for comparison, we also included the performance of 2/3/4-bit Q-ViT in Table \ref{tab:coco}. Our method outperform BiBERT, RBONN, and Bi-Real Net in terms of AP@[.5,.95] by 11.8\%, 11.2\%, and 10.8\% when using the Mask R-CNN framework with the Swin-Tiny backbone.

Furthermore, our Bi-ViT model showed superior performance on other APs with different IoU thresholds and achieved 9.5\% lower AP than the 2-bit Q-ViT model. Our method also yielded a 1-bit transformer-based detector with a performance of 23.0\% AP lower than the real-valued counterpart (43.7\% vs. 20.7\%) while utilizing 2.3$\times$ less memory.
When using the Cascade Mask R-CNN framework with the Swin-Tiny backbone, our method achieved 29.0\% AP@[.5,.95], outperforming BiBERT, RBONN, and Bi-Real Net by 12.3\%, 11.5\%, and 10.9\% mAP, respectively.

In conclusion, our method demonstrated superior performance in terms of AP with various IoU thresholds and AP for objects of different sizes on the COCO dataset, showing its applicability and superiority in various application settings compared to previous binary neural networks.

\section{Conclusion}
In this paper, we present Bi-ViT, an improved version of fully-binarized ViTs that offers a high compression ratio and acceptable performance. Initially, we establish a empirical framework for fully-binarized ViT and analyze the bottlenecks of the baseline. Our empirical analysis shows that attention distortion in MHSA is the primary cause of the significant drop in ViT binarization, which results from gradient vanishing and ranking disorder. 
To address these issues, we introduce a learnable scaling factor that reactivates vanished gradients, which we illustrate through both theoretical and experimental analysis. Additionally, we propose ranking-aware distillation for Bi-ViT, which rectifies disordered ranking in a teacher-student framework. Our work provides a comprehensive analysis and effective solutions for the crucial issues in ViT full binarization, paving the way for the extreme compression of ViT.

{\small
\bibliographystyle{ieee_fullname}
\bibliography{egbib}
}

\end{document}